\documentclass[sigconf]{acmart}

\usepackage{epsfig}
\usepackage{graphicx}
\usepackage{amsmath}

\usepackage{amssymb}
\usepackage{booktabs}
\usepackage{multirow}
\usepackage{bigstrut}
\usepackage{makecell}
\usepackage{ulem}
\usepackage[labelformat=simple]{subcaption}

\newcommand{\yuqi}[1]{\textcolor{black}{#1}}
\newcommand{\new}[1]{\textcolor{black}{#1}}
\AtBeginDocument{%
  }

\copyrightyear{2023} \acmYear{2023} \setcopyright{acmlicensed}\acmConference[MM '23]{Proceedings of the 31st ACM International Conference on Multimedia}{October 29-November 3, 2023}{Ottawa, ON, Canada} \acmBooktitle{Proceedings of the 31st ACM International Conference on Multimedia (MM '23), October 29-November 3, 2023, Ottawa, ON, Canada} \acmPrice{15.00} \acmDOI{10.1145/3581783.3612228} \acmISBN{979-8-4007-0108-5/23/10}

\begin{document}

\title{Recurrent Self-Supervised Video Denoising with Denser Receptive Field}

\author{Zichun Wang}
\authornote{Both authors contributed equally to this research.}
\email{xiaoding310@gmail.com}
\affiliation{%
  \institution{Beijing Institute of Technology}
}

\author{Yulun Zhang}
\email{yulun100@gmail.com}
\authornotemark[1]
\affiliation{%
  \institution{ETH Z\"{u}rich}
}

\author{Debing Zhang}
\email{debingzhangchina@gmail.com}
\affiliation{%
  \institution{Xiaohongshu Inc}
}

\author{Ying Fu}
\authornote{Corresponding author.}
\email{fuying@bit.edu.cn}
\affiliation{%
  \institution{Beijing Institute of Technology}
}

\renewcommand{\shortauthors}{Wang et al.}


\begin{abstract}

Self-supervised video denoising has seen decent progress through the use of blind spot networks.
However, under their blind spot constraints,
previous self-supervised video denoising methods suffer from significant information loss and texture destruction in either the whole reference frame or neighbor frames, due to their inadequate consideration of the receptive field. Moreover, 
the limited number of available neighbor frames
in previous methods
leads to 
the discarding of distant temporal information.
Nonetheless, simply adopting existing recurrent frameworks does not work, since they easily break the constraints on the receptive field imposed by self-supervision. In this paper, we propose RDRF
for self-supervised video denoising, which not only fully exploits both the reference and neighbor frames with a denser receptive field, but also better leverages the temporal information from both local and distant neighbor features. First, towards a comprehensive utilization of information from both reference and neighbor frames, RDRF realizes a denser receptive field by taking more neighbor pixels along the spatial and temporal dimensions. Second, it features a self-supervised recurrent video denoising framework, which concurrently integrates distant and near-neighbor temporal features.
This enables long-term bidirectional information aggregation, 
while mitigating error accumulation in the plain recurrent framework. 
Our method exhibits superior performance on both synthetic and real video denoising datasets.
Codes will be available at \url{https://github.com/Wang-XIaoDingdd/RDRF}.
\end{abstract}



%

\maketitle

\begin{figure}[t]
\centering
\includegraphics[width=0.98\linewidth]{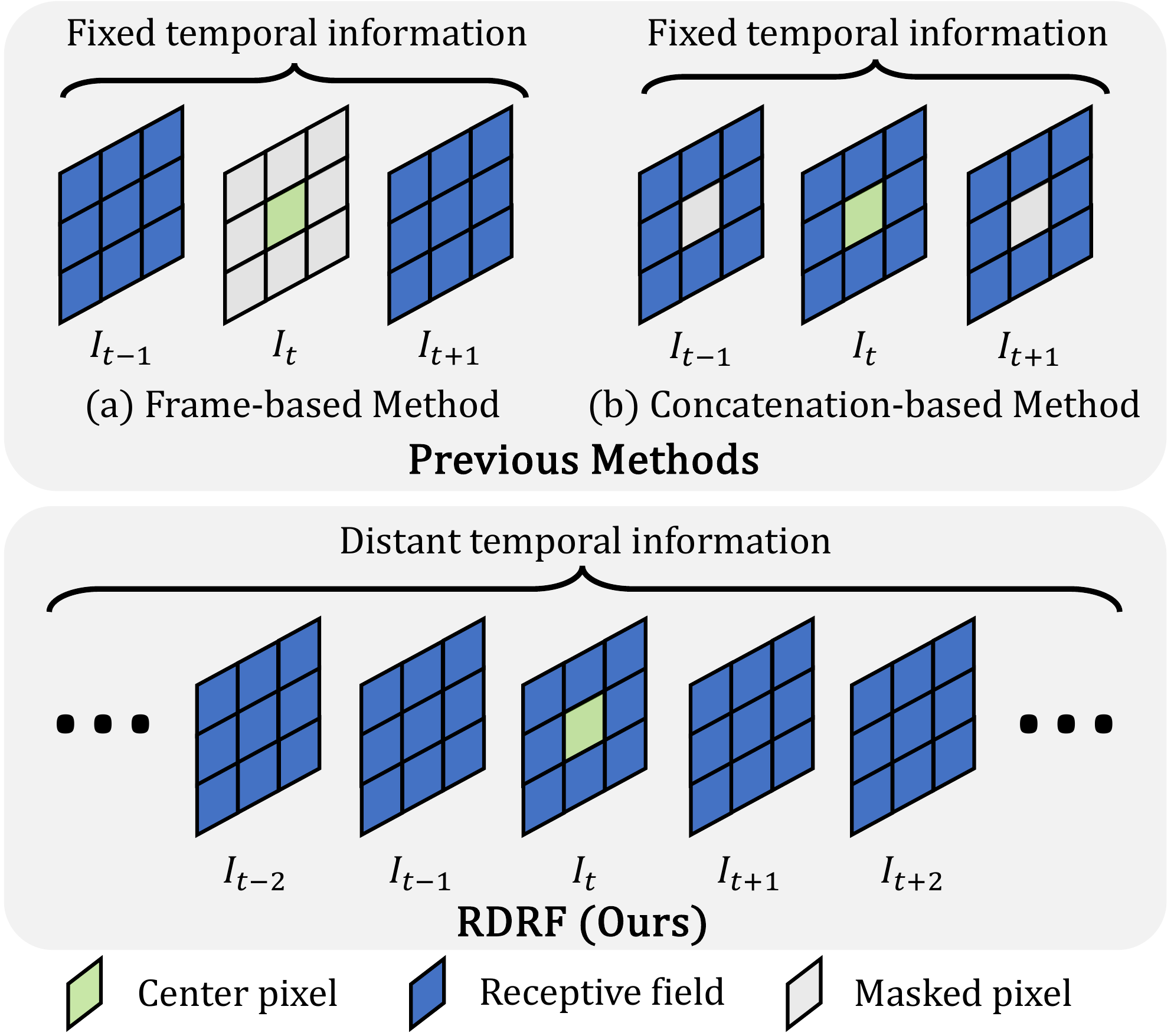}
\vspace{-2mm}
\caption{
\textbf{Receptive field comparison.} 
(a) Frame-based methods \cite{dewil2021self,yu2020joint,li2021learning} use neighbor frames only to predict the reference frame, and the reference frame is fully masked. (b) Concatenation-based method \cite{sheth2021unsupervised} treats the reference and neighbor frames in the same way, where the restricted receptive field for neighbors leads to under-utilization of temporal information. 
(c) RDRF realizes a denser receptive field of both reference and neighbor information.
Besides, it adopts the recurrent framework under
the blind spot requirement, which takes advantage of distant
temporal information.
}
\vspace{-4mm}
\label{fig:teaser}
\end{figure}

\vspace{-2mm}
\section{Introduction}

Video denoising is a fundamental topic in low-level computer vision tasks \cite{tassano2019dvdnet,tassano2020fastdvdnet,ehret2019model,xue2019video,qi2022real,mehta2021evrnet}. Specifically, noise can greatly degrade the visual quality of videos, especially under \yuqi{light-deficient conditions}. 
\yuqi{Besides, the poor visual quality of noisy videos} will also adversely impact downstream computer vision tasks \cite{10.1145/3126686.3126727, deng2022nightlab}.

Recently, thanks to powerful deep neural networks, supervised video denoising
\cite{tassano2019dvdnet, tassano2020fastdvdnet, li2022unidirectional,maggioni2021efficient,fu2022low,vaksman2021patch} have achieved impressive performance. Regrettably, 
the success of supervised video denoising methods often relies on large-scale paired noisy-clean datasets. Nonetheless, collecting high-quality paired videos covering various types of sensors and noise levels can be extremely labor-intensive. Furthermore, paired videos are almost unavailable in some complex scenarios, \textit{e.g.}, microscopy videos and dynamic scene videos.
\yuqi{U}nlike image denoising datasets \cite{abdelhamed2018high,plotz2017benchmarking}, where clean images are 
collected based on either the long exposure time or averaging multiple noisy images, 
collecting videos can result in motion blur in dynamic scenes, which severely degrades 
their quality.
Consequently, existing 
paired 
video denoising datasets are often captured under degraded or simplified settings \cite{fu2022low,chen2019seeing,yue2020supervised}, 
whose generalization ability to real world situations can be greatly constrained \cite{fu2022low}.

\yuqi{Therefore}, to eliminate the dependence on large-scale paired datasets, several attempts have been made for self-supervised video denoising. Inspired by either the pioneer work Noise2Noise \cite{lehtinen2018noise2noise} or the subsequent blind spot networks \cite{batson2019noise2self,abdelhamed2018high,krull2019noise2void,laine2019high},
previous studies can be divided into two categories, \textit{i.e.}, frame-based \cite{abdelhamed2018high,batson2019noise2self,krull2019noise2void,dewil2021self} and concatenation-based \cite{sheth2021unsupervised} methods. However, their performance is limited by the insufficient receptive field on either the reference frame or the neighbor frames, as shown in Figure \ref{fig:teaser}.

First, frame-based methods attempt to apply Noise2Noise \cite{lehtinen2018noise2noise} framework on videos.
These methods 
align the consecutive frames using the estimated optical flow. Then, under the assumption that the warped frames are well-aligned,
it adopts the Noise2Noise \cite{lehtinen2018noise2noise} strategy
for self-supervised denoising. 
However, they 
are susceptible to the
error resulting from inaccurate optical flow estimation, where the basic assumption,
\textit{i.e.}, the same underlying clean signal,
is no longer satisfied.
Moreover, frame-based methods require reference frames to be masked to avoid learning identity mapping. This dramatically reduces the receptive field of these methods, 
where the useful information in the reference frame is greatly lost while restoring the reference frame itself. 
Second, inspired by the image blind spot networks \cite{abdelhamed2018high,batson2019noise2self,krull2019noise2void}, another line of work focuses on designing video blind spot networks. Recently, UDVD \cite{sheth2021unsupervised} extends the image blind-spot network \cite{laine2019high} based on a video denoising network \cite{tassano2020fastdvdnet}.
However, UDVD \cite{sheth2021unsupervised} applies the same blind spot processing to both the reference frame and neighboring frames, while this leads to a significant loss of neighbor frame information.

Apart from the limited receptive field, another challenge for existing methods is to fully utilize the temporal information. While long-range temporal features
have been proven to be crucial for video restoration \cite{chan2021basicvsr,chan2022basicvsr++,haris2019recurrent},
\new{current short-term 
temporal
aggregation 
is insufficient, particularly for long-sequence videos. 
This limits the availability of 
long-term temporal
information, 
as it only considers local neighbor frames within a very limited window size, which typically spans three or five contiguous frames \cite{sheth2021unsupervised,dewil2021self}.
And simply enlarging the window size can 
bring challenges to the balance of efficiency and effectiveness. 
}
However, it cannot be solved by naively adopting existing recurrent fashion, since it breaks the blind spot requirement and thus fails to recover
the clean scenes.

In this paper, we propose RDRF, a \yuqi{self-supervised video denoising} method to tackle these issues. 
First, under a fine consideration of the blind spot requirement for self-supervision, RDRF realizes a denser receptive field. This allows for fully exploiting the information that is existed both in the reference frame and neighbor frames, providing valuable information for improved detail reconstruction.
\new{Second, under the blind spot constraint, RDRF adopts the recurrent fashion.
It significantly increases the usable  temporal information, 
leveraging more inter-frame similarity to enhance the restoration process.} Specifically, our method contains two parallel branches, focusing on the fusion of distant and near-neighbor temporal information respectively.
For the long-range branch, we propose a 
Distant Feature Fusion module, 
which combines the reference blind spot feature with the temporally propagated non-blind feature in a spatially-adaptive way.
For the local branch, we propose a Local Feature Extraction module, which utilizes 3D blind spot convolution to fully utilize the spatial and temporal local information, and also eases the error accumulation in the plain recurrent framework. By our design, the recovery
can be aided by both distant and near neighbor frames with comprehensive exploitation.

We sum up our contributions in the following:

\begin{itemize}
	
	\item We present a novel method for self-supervised video denoising, which fully exploits both reference and neighbor frames under blind spot requirement, enabling a denser receptive field for comprehensive information utilization.

	\item Towards full exploitation of temporal features, our method recurrently integrates distant and near-neighbor temporal features, which realizes robust long-term information aggregation and error accumulation mitigation concurrently.
	
	\item Experiments results show that our method achieves state-of-the-art performance on various synthetic and real world video denoising datasets.

\end{itemize}

\section{Related Work}

In this section, we review relevant studies on supervised video denoising, unsupervised
image and video denoising.

\vspace{-2mm}

\subsection{Supervised Video Denoising}

Natural videos exhibit significant redundancy along the temporal dimension, which requires video denoising methods to fully utilize the temporal information compared to processing a single image only.
Traditional methods group spatially and temporally similar patches \cite{buades2016patch,maggioni2012video}, but the searching process can be time-consuming, and the performance is sometimes unsatisfactory. Recently, deep learning based methods have shown better performance by employing a sliding window scheme to aggregate temporal information. 
ViDeNN \cite{claus2019videnn} uses two CNN-based fusion networks, one for processing spatial and the other for temporal information. In order to better utilize temporal information to assist in the recovery of the current frame, some studies have introduced various components to advance the fusion process,  \textit{e.g.}, optical flow \cite{xue2019video,tassano2019dvdnet}, deformable convolution \cite{yue2020supervised}, cascaded U-Net \cite{tassano2020fastdvdnet}, non-local patch grouping \cite{vaksman2021patch}, kernel prediction network \cite{mildenhall2018burst,xia2020basis,xu2020learning} and channel shifting \cite{rong2020burst}. 
\new{However, these methods 
may discard the distant temporal information, where only one or two frames before and after the reference frame are considered.}
Recently, the recurrent video denoising framework has attracted increasing attention owing to its capability to capture long-term temporal information. 
These methods typically rely on the bidirectional or unidirectional temporal feature, featuring guided deformable attention \cite{liang2022recurrent,chan2022generalization}, mimicked backward recurrent module \cite{li2022unidirectional}, and low computational costs \cite{xiang2022remonet,ostrowski2022bp,maggioni2021efficient}.
All of these methods are designed for supervised denoising, and cannot work when only noisy observations are provided.

\begin{figure*}[htbp]
  \centering
  \includegraphics[width=1\linewidth]{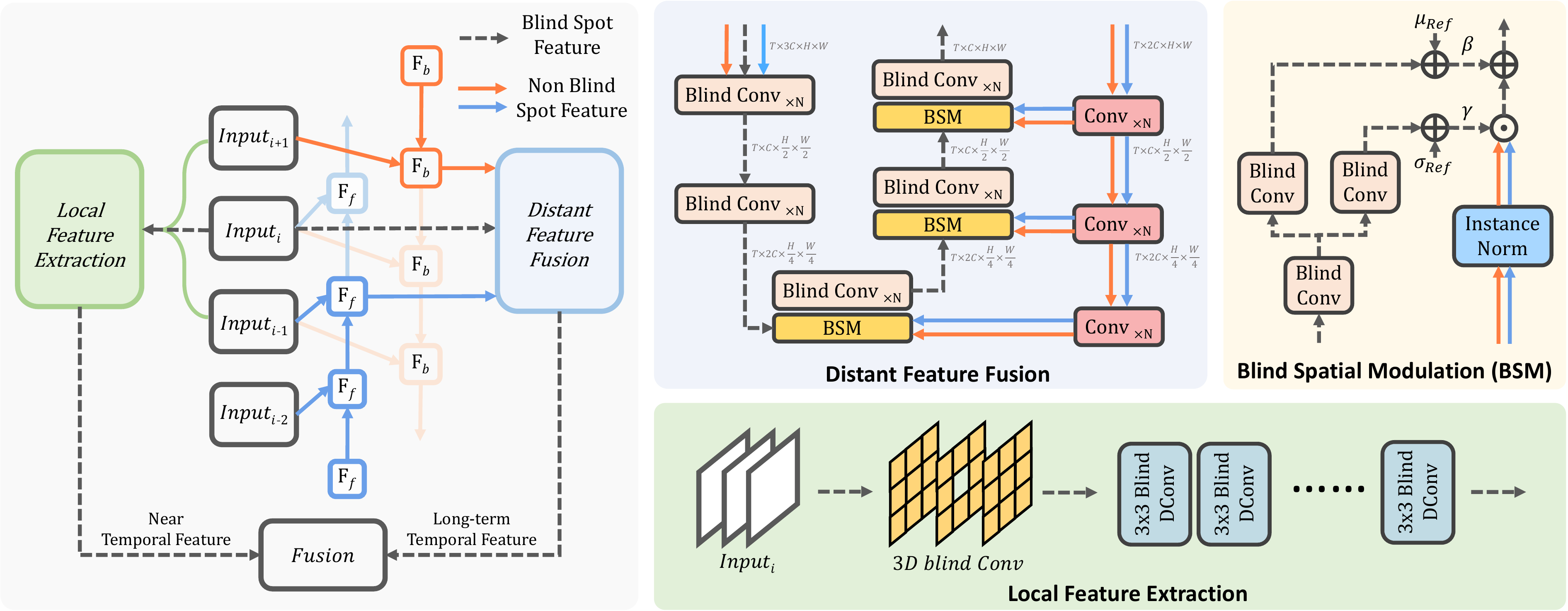}
  \vspace{-6mm}
  \caption{Illustration of the general structure of RDRF. Our approach comprises two branches, each with the specific goal,
  \textit{e.g.},
  local textures and long-term information.
  Finally, the outputs from both branches are combined to produce the final result. 
  }
  
  \label{fig:overall_arch}
\vspace{-2mm}
\end{figure*}

\subsection{Unsupervised Image Denoising}

Under situations where paired noisy-clean images are unavailable, researchers seek to exploit the internal information in noisy images. These methods are trained on different types of data, including pseudo noisy-clean pairs, noisy-noisy pairs, and noisy data only. For methods aiming to generate pseudo-noisy-clean pairs, GCBD \cite{chen2018image} synthesizes realistic noisy images by a generative adversarial network (GAN), which is then advanced  by introducing extra knowledge of a self-supervised network \cite{wu2020unpaired}, or taking more noise components into account \cite{jang2021c2n}. 
Besides, inspired by the semi-supervised method Noise2Noise \cite{lehtinen2018noise2noise}, some methods are trained on noisy-noisy pairs. To synthesize noisy image pairs, several techniques have been proposed, \textit{e.g.},
sub-sampling from original images \cite{huang2021neighbor2neighbor}, generating noisier-noisy pairs \cite{moran2020noisier2noise,xu2020noisy}, or by a data augmentation technique \cite{pang2021recorrupted}. 
Some other methods are trained directly on noisy images. Noise2Void \cite{krull2019noise2void} and Noise2Self \cite{batson2019noise2self} initially propose the blind spot network (BSN), while the performance and efficiency are unsatisfying. To this end, Laine19 \cite{laine2019high}, D-BSN \cite{wu2020unpaired} and Blind2Unblind \cite{wang2022blind2unblind} are further proposed with advanced BSN designs and better performance.
Recently, the application of BSN is extended from spatially independent noise to spatially dependent real noise \cite{neshatavar2022cvf,lee2022ap,wang2023lg}. However, these methods still focus on denoising a single image, and the temporal information can not be utilized.

\subsection{Unsupervised Video Denoising}

Based on the method used, current self-supervised video denoising methods can be divided into two main categories: those based on Noise2Noise (N2N) \cite{lehtinen2018noise2noise} or the blind spot networks \cite{krull2019noise2void,laine2019high,batson2019noise2self}.  
The first type of method follows the assumption of N2N, which requires noisy observations under the same scene.
For videos, 
adjacent frames can be seen as under the same scene with motion transformation. 
Thus, under this assumption, Frame2Frame (F2F) \cite{ehret2019model} first warps adjacent frames
based on the TV-L1 optical flow \cite{zach2007duality}, then 
follows N2N
for denoising. This is further enhanced with a trainable optical flow estimator \cite{yu2020joint} for more accurate alignment.
Nonetheless, the adopted image-based denoiser is not suitable for 
video denoising task. 
Therefore, Multi-Frame2Frame \cite{dewil2021self} takes multiple neighbor frames as input to predict the masked reference frame.
A twin sampler strategy is also proposed to 
prevent from learning the identity mapping \cite{li2021learning}.
However, they all suffer from inaccurate motion and occlusion estimation, especially when the motion is relatively high or intensity conversion are not attributed to the translation. Moreover, they satisfy the self-supervision requirement at the cost of discarding the reference frame. The limited useful information significantly constrains their performance.

Another type of method try to extend the image blind spot network to videos, where the reference frame information can thus be utilized. Specifically, 
based on the architecture of FastDVDnet \cite{tassano2020fastdvdnet}, UDVD \cite{sheth2021unsupervised} takes the stack of frames into a blind spot network, while the supporting neighbor features are processed 
under a constrained receptive field, resulting in destroyed temporal similar patterns due to their course consideration of the self-supervision requirement for videos. 
Also, current methods with short-term temporal aggregation fall short of valuable distant temporal information.
This limits the availability of long-term similar patterns, as it only considers frames
within limited window size, which typically operates on only three or five contiguous frames.

\section{Method}

In this section, we first go into details about the motivation, and then present detailed demonstrations of our 
two core designs, \textit{i.e.}, Distant Feature Fusion and Local Feature Extraction.

\subsection{Motivation and Modeling}

Despite the promising performance achieved by image denoising methods, video denoising
intrinsically 
differs
due to the huge amount of extra information along the temporal dimension \cite{fu2022low,tassano2020fastdvdnet,tassano2019dvdnet,xue2019video}. 
The recovery of a video frame requires two types of information: the reference frame provides one perfectly aligned noisy observation of the underlying scene, while its misaligned neighbor frames offer abundant similar patterns for recovery.  Therefore, designing a video denoising method necessitates the full exploitation of both the reference and neighbor frames.

\noindent \textbf{Full exploitation for reference and neighbor frames.} However, for self-supervised denoising, it is not feasible to simply increase the density of receptive fields and utilize all information, which violates the blind spot constraint \cite{krull2019noise2void,laine2019high,batson2019noise2self} and leads to learning identity mapping. 
Therefore, the key lies in the careful consideration to maximize the receptive field, while respecting the blind spot requirement. Nonetheless, the coarse consideration of such constraint inevitably brings information loss and degraded performance for current methods.
Some methods solely rely on the information from neighbor frames \cite{dewil2021self,sheth2021unsupervised,ehret2019model,ehret2019model,yu2020joint,li2021learning}, and the missing of the whole reference frame 
means the lack of a perfectly aligned noisy observation from the reference frame, leading to inferior performance.
Other methods suffer from  the destruction of useful hints in neighbor frames \cite{sheth2021unsupervised}, since the reference and neighbor frames are processed in the same way following the blind spot requirement. 
The restricted receptive field originates from the under-consideration of the blind spot requirement.
Instead, our observation is that, the blind spot for the reference frame is already enough for self-supervision. Instead, neighbor frames are not limited by such constraints, which it is usually omitted by previous studies, including the previous SOTA method UDVD \cite{sheth2021unsupervised}.

\noindent \textbf{Gathering long-term temporal information.} Another challenge for previous methods is to effectively aggregate temporal information, especially for long-sequence videos.
First, just as the spatial non-local similarity has been proven indispensable for image restoration \cite{buades2005non,liang2021swinir}, 
leveraging more long-term dependencies along the temporal dimension is also the key to video denoising \cite{maggioni2012video,vaksman2021patch,fu2022low}. 
\new{Unfortunately, previous methods are unable to accumulate distant temporal similarities, while using a large window size brings challenges for efficiency.}
Second, for distant temporal information aggregation, simply adopting a recurrent framework is also not an optimal choice. Recent studies have revealed error accumulation is inevitable due to the propagation mechanism \cite{kang2022error,zhou2022revisiting,huang2022neural}, and relying on temporally propagated features only is not enough for effective exploitation of the redundant neighbor patterns.

\noindent\textbf{Our method.} We aim to address both of these issues in this paper. First, based on different blind spot requirements for reference and neighbor frames, we treat them in a blind and non-blind manner respectively. 
Second, toward full utilization of distant temporal information, at the same time easing error accumulation, we integrate both long-term and near-neighbor features in a concurrent way.

\noindent \textbf{Problem modeling.} Given a noisy video sequence of size $x\in\mathbb{R}^{T \times H \times W \times C}$, where $C$, $T$, $H$ and $W$ are the channel, video length, height, and width respectively.
In Figure \ref{fig:overall_arch}, we provide the architecture of RDRF, which contains two branches in parallel. For long-term temporal information
aggregation, 
neighbor 
temporal 
features are first propagated forward and backward respectively without the blind spot requirement, which prevents
the loss of temporal similar patterns in neighbor frames. 
The propagated long-term temporal features are then fused with the blind feature of the reference frame by the Distant Feature Fusion. 
For the near-temporal feature, Local \vspace{-0.078mm}Feature Extraction extracts the detailed structure within the temporally adjacent frames.
Finally, the output of these two parallel branches is fused together as the final output.

\subsection{Distant Feature Fusion}

To address the severely limited receptive field
and the 
insufficient
long-term temporal features,
we redesign the main components for self-supervised video denoising: propagation and fusion. 

\noindent \textbf{Temporal information propagation.} Propagation decides the way we leverage the information that existed in a video sequence. Previous methods adopt local propagation \cite{sheth2021unsupervised,dewil2021self,ehret2019model}, 
\new{where the restoration can only be aided by 
the short-term temporal information.}
Instead, 
we take advantage of the bidirectional propagation scheme, where temporal information can be propagated both forward and backward
\new{for long-term temporal patterns.}
However, current supervised recurrent frameworks \cite{chan2021basicvsr,chan2022basicvsr++}
for $I^{Clean}_{i}$ require $I^{Noisy}_{i}$ as input, which breaks the requirement for self-supervision. 
A naive solution is to
process neighbor frames ‘blindly’, in the same way as the current frame \cite{sheth2021unsupervised}, while it results in the texture destruction in neighbor frame features. 
Instead, as shown in Figure \ref{fig:overall_arch}, we first propagate the long-term information in a non-blind manner, where neighbor features can be fully extracted. Then, for a noisy input $x_i$, its neighboring frames $x_{i-1}$ and $x_{i+1}$, the forward and backward propagated features denoted as $h_{i-1}^f$ and $h_{i+1}^b$, the temporal propagation is computed as:
\vspace{-1mm}
\begin{equation}
\begin{aligned}
& h_i^b=F_b\left(x_{i+1}, h_{i+1}^b\right) \\
& h_i^f=F_f\left(x_{i-1}, h_{i-1}^f\right),
\vspace{-4mm}
\end{aligned}
\end{equation}
where $F_b$ and $F_f$ denote the backward and forward propagation modules respectively. 
We implement 
$F_b$ and $F_f$
with Restormer \cite{zamir2022restormer} blocks
for enhanced long-distance dependencies modeling.
Here, note that the reference frame $x_i$ itself will not be taken as the input of $F_b$ and $F_f$
for
the blind spot requirement. 
In other words, only neighbor frames are processed in a "non-blind" manner, enabling long-term neighbor feature aggregation with a denser receptive field. 
By doing so, we are also free of the special structure of blind spot networks for $F_b$ and $F_f$, allowing us to benefit from the latest design of the supervised method, \textit{e.g.}, Restormer block \cite{zamir2022restormer}.

\noindent \textbf{Fusion of reference and neighbor information.} Next, after obtaining fully extracted temporal features from neighbor frames, the next step is to effectively fuse the information from the reference frame and the propagated features. 
Different from existing methods that directly concatenate reference and neighbor features, then process them through the blind spot network \cite{sheth2021unsupervised}, we benefit from the 
non-blind neighbor features
to better assist the recovery of the current frame in a spatially adaptive way, inspired by previous works \cite{park2019semantic,lu2021masa}. 
Specifically, our Distant Feature Fusion module takes two sets of features as inputs. 
A blind spot network \cite{laine2019high} first takes the neighbor features and temporal propagated features, which employ four branches with limited receptive fields in distinct directions. To produce the blind spot, each branch applies a shift of a single pixel to the features.
Following the implementation of 
this architecture, our Blind Spatial Modulation module remaps the distribution of temporal features to the reference frame.  
To generate the parameters $\beta$ and $\gamma$, the reference feature is first concatenated with temporal features and fed into the blind convolutions following HQ-SSL \cite{laine2019high}. Then we apply instance normalization on the temporal features as:
\begin{equation}
{Nei}_{i}^c \longleftarrow \frac{{Nei}_{i}^c-{\mu}_{i}^c}{{\sigma}_{i}^c},
\end{equation}
where ${Nei}_{i}^c$ is the concatenation of temporal propagated feature $h_i^b$ from backward and $h_i^f$ from forward at $i^{th}$ frame in channel $c$. ${\sigma}_{i}^c$ and ${\mu}_{i}^c$ is the standard deviation and mean for ${Nei}_{i}^c$ as:
\begin{equation}
\begin{gathered}
{\mu}_{i}^c=\frac{1}{H W} \sum_{h, w} {Nei}_{i}^{c, h, w}, \\
{\sigma}_{i}^c=\sqrt{\frac{1}{H W} \sum_{h, w}\left({Nei}_{i}^{c, h, w}-{\mu}_{i}^c\right)^2}.
\end{gathered}
\end{equation}
where $H$ and $W$ are the height and width of ${Nei}_{i}^c$. Then, 
based on 
the reference frame statistic, we update $\beta$ and $\gamma$ as:
\begin{equation}
\begin{aligned}
& {\beta} \longleftarrow {\beta}+\hat{{\mu}}_{i}^c, \\
& {\gamma} \longleftarrow {\gamma}+\hat{{\sigma}}_{i}^{c},
\end{aligned}
\end{equation}
where $\hat{{\mu}}_{i}^c$ and $\hat{{\sigma}}_{i}^{c}$ denotes the mean and standard deviation of reference feature. Finally, we can remap the distribution for the non-blind temporal neighbor feature as follows:
\begin{equation}
{F}_{R e f} \longleftarrow {F}_{R e f} \cdot {\gamma}+{\beta}.
\end{equation}
The neighbor features are re-modulated in a spatially adaptive way, conditioned on the scene difference between the frames.
This is then processed by the blind convolution for the overall output.

\begin{table*}[t]
  \centering
\setlength{\tabcolsep}{2mm}{
 \renewcommand{\arraystretch}{1}{

\begin{tabular}{cccccccccc}
\toprule
Set8  & \multicolumn{2}{c}{Traditional} & \multicolumn{4}{c}{Supervised} & \multicolumn{3}{c}{Unsupervised} \\

sigma & \multicolumn{1}{c}{\begin{tabular}[c]{@{}c@{}}VBM4D \\\cite{maggioni2012video}\end{tabular}}&
\multicolumn{1}{c}{\begin{tabular}[c]{@{}c@{}}VNLB \\ \cite{lebrun2013nonlocal}\end{tabular}}&
\multicolumn{1}{c}{\begin{tabular}[c]{@{}c@{}}VNLnet \\ \cite{davy2019non}\end{tabular}}&
\multicolumn{1}{c}{\begin{tabular}[c]{@{}c@{}}DVDnet \\ \cite{tassano2019dvdnet}\end{tabular}}&
\multicolumn{1}{c}{\begin{tabular}[c]{@{}c@{}}FastDVDnet \\ \cite{tassano2020fastdvdnet}\end{tabular}}&
\multicolumn{1}{c}{\begin{tabular}[c]{@{}c@{}}FloRNN \\ \cite{li2022unidirectional}\end{tabular}}&
\multicolumn{1}{c}{\begin{tabular}[c]{@{}c@{}}MF2F \\ \cite{dewil2021self}\end{tabular}}&
\multicolumn{1}{c}{\begin{tabular}[c]{@{}c@{}}UDVD \\ \cite{sheth2021unsupervised} \end{tabular}}&
\multicolumn{1}{c}{\begin{tabular}[c]{@{}c@{}}RDRF \\(Ours) \end{tabular}} \\

\cmidrule[0.4pt](lr){1-1} 
\cmidrule[0.4pt](lr){2-3} 
\cmidrule[0.4pt](lr){4-7} 
\cmidrule[0.4pt](lr){8-10} 

    10    & 36.05/- & 37.26/- & 37.28/.9606 & 36.08/.9510 & 36.44/.9540 & \uline{37.57/.9639} & 36.01/.9379 & 36.36/.9510 & \textbf{36.67/.9547} \\
    20    & 32.19/- & 33.72/- & 34.02/.9273 & 33.49/.9182 & 33.43/.9196 & \uline{34.67/.9379} & 33.79/.9115 & 33.53/.9167 & \textbf{34.00/.9251} \\
    30    & 30.00/- & 31.74/- & -     & 31.68/.8862 & 31.68/.8889 & \uline{32.97/.9138} & 32.20/.8831 & 31.88/.8865 & \textbf{32.39/.8978} \\
    40    & 28.48/- & 30.39/- & 30.72/.8622 & 30.46/.8564 & 30.46/.8608 & \uline{31.75/.8911} & 30.64/.8413 & 30.72/.8595 & \textbf{31.23/.8725} \\
    50    & 27.33/- & 29.24/- & -     & 29.53/.8289 & 29.53/.8351 & \uline{30.80/.8696} & 28.90/.7775 & 29.81/.8349 & \textbf{30.31/.8490} \\
\midrule
    avg   & 30.81/- & 32.47/- & -     & 32.29/.8881 & 32.31/.8917 & \uline{33.55/.9153} & 32.31/.8703 & 32.46/.8897 & \textbf{32.92/.8998} \\

\bottomrule
\end{tabular}%

}
}
  \caption{
  Performance comparison on Set8 \cite{tassano2020fastdvdnet} dataset. 
  The best results for un-/supervised methods are in bold and underlined.}
  \vspace{-5mm}
  \label{tab:res_set8}
\end{table*}

\begin{table*}
  \centering
\setlength{\tabcolsep}{2mm}{
 \renewcommand{\arraystretch}{1}{

\begin{tabular}{cccccccccc}
\toprule
Davis  & \multicolumn{2}{c}{Traditional} & \multicolumn{4}{c}{Supervised} & \multicolumn{3}{c}{Unsupervised} \\

sigma & \multicolumn{1}{c}{\begin{tabular}[c]{@{}c@{}}VBM4D \\\cite{maggioni2012video}\end{tabular}}&
\multicolumn{1}{c}{\begin{tabular}[c]{@{}c@{}}VNLB \\ \cite{lebrun2013nonlocal}\end{tabular}}&
\multicolumn{1}{c}{\begin{tabular}[c]{@{}c@{}}VNLnet \\ \cite{davy2019non}\end{tabular}}&
\multicolumn{1}{c}{\begin{tabular}[c]{@{}c@{}}DVDnet \\ \cite{tassano2019dvdnet}\end{tabular}}&
\multicolumn{1}{c}{\begin{tabular}[c]{@{}c@{}}FastDVDnet \\ \cite{tassano2020fastdvdnet}\end{tabular}}&
\multicolumn{1}{c}{\begin{tabular}[c]{@{}c@{}}FloRNN \\ \cite{li2022unidirectional}\end{tabular}}&
\multicolumn{1}{c}{\begin{tabular}[c]{@{}c@{}}MF2F \\ \cite{dewil2021self}\end{tabular}}&
\multicolumn{1}{c}{\begin{tabular}[c]{@{}c@{}}UDVD \\ \cite{sheth2021unsupervised} \end{tabular}}&
\multicolumn{1}{c}{\begin{tabular}[c]{@{}c@{}}RDRF \\(Ours) \end{tabular}} \\

\cmidrule[0.4pt](lr){1-1} 
\cmidrule[0.4pt](lr){2-3} 
\cmidrule[0.4pt](lr){4-7} 
\cmidrule[0.4pt](lr){8-10} 

10    & 37.58/- & 38.85/- & 39.56/.9707 & 38.13/.9657 & 38.71/.9672 & \uline{40.16/.9755} & 38.04/.9566 & 39.17/.9700 & \textbf{39.54/.9717} \\
20    & 33.88/- & 35.68/- & 36.53/.9464 & 35.70/.9422 & 35.77/.9405 & \uline{37.52/.9564} & 35.61/.9359 & 35.94/.9428 & \textbf{36.40/.9473} \\
30    & 31.65/- & 33.73/- & -     & 34.08/.9188 & 34.04/.9167 & \uline{35.89/.9440} & 33.65/.9065 & 34.09/.9178 & \textbf{34.55/.9245} \\
40    & 30.05/- & 32.32/- & 33.32/.8996 & 32.86/.8962 & 32.82/.8949 & \uline{34.66/.9286} & 31.50/.8523 & 32.79/.8949 & \textbf{33.23/.9032} \\
50    & 28.80/- & 31.13/- & -     & 31.85/.8745 & 31.86/.8747 & \uline{33.67/.9131} & 29.39/.7843 & 31.80/.8739 & \textbf{32.20/.8832} \\
\midrule
avg   & 32.39/- & 34.34/- & -     & 34.52/.9195 & 34.64/.9188 & \uline{36.38/.9435} & 33.64/.8871 & 34.76/.9199 & \textbf{35.18/.9260} \\
\bottomrule
\end{tabular}%

}
}
  \caption{Performance comparison on Davis \cite{pont20172017} dataset. The best results for un-/supervised methods are in bold and underlined.
  }
  \vspace{-6mm}
  \label{tab:res_davis}
\end{table*}

\subsection{Local Feature Extraction}

The recurrent network can naturally aggregate long-term temporal information, bringing abundant similar patterns 
\new{from its distant neighbor frames.}
However, errors generated in each time step can be accumulated meanwhile \cite{kang2022error,zhou2022revisiting,huang2022neural}, which limits the effectiveness of auxiliary neighbor features. Thus, relying on the propagated neighbor features is not enough
for recovery.

In addition, the closer to the reference frame, the more similar the content of neighbor frames. These neighbor frames typically contain the most useful information, which should be carefully considered.
Thus, besides capturing the temporally long-term information, we further propose the Local Feature Extraction module to better focus on the near neighbor feature.

Specifically, the main goal of blind spot networks is to ensure that the output at each position is not influenced by the input.
Under such requirement, we fully utilize both spatially and temporally neighbor pixels to recover the center masked pixel. As shown in Figure \ref{fig:overall_arch}, we propose a blind 3D convolution for near-neighbor spatial-temporal features extraction, which is defined as:
\begin{equation}
{Out}_i^c={\begin{bmatrix}x_{i-1}, x_{i}, x_{i+1}\end{bmatrix}} \otimes \left({Wei}_i^c \circ {Mask}\right),
\end{equation}
where ${Out}_i^c$ is the output $c^{th}$ feature of the 3D blind convolution, ${Wei}_i^c$ and $Mask$ denote the $3\times3\times3$ kernel weight and corresponding binary mask $m$, $\otimes$ and $\circ$ denote convolution and element-wise product operator.
Note that the center of the kernel weight is set to zero, in order to meet the self-supervised blind spot constraint.
Then, inspired by the previous self-supervised image denoising method \cite{wu2020unpaired},  
we maintain the blind spot requirement by stacking dilated convolution with a stride of 2 to extract deep local features. 

Finally, after obtaining the near-neighbor and distant temporal features, we use $1\times1$ convolutions to gradually transform the high-dimensional feature into the final output.

\section{Experiments}
\vspace{-1mm}

In this section, we introduce datasets and setup details, providing the performance comparison and corresponding analysis.

\vspace{-2mm}

\subsection{Dataset and Setup Details}

\noindent \textbf{Natural videos.} 
We train our method with the widely used DAVIS train set \cite{pont20172017} for the removal of synthetic Gaussian noise,
and conduct experiments with natural videos by introducing independent and identically distributed  Gaussian noise.
We evaluate 
on two datasets,
including Davis test set \cite{pont20172017} and Set8 \cite{tassano2020fastdvdnet} dataset.
Four videos from the Derfs Test Media collection \cite{dewil2021self} and four GoPro-shot footage make up the Set8 dataset.
To maintain consistency with previous settings, we use initial 85 frames in the Derfs dataset.

\noindent \textbf{Raw videos.} 
The CRVD dataset \cite{yue2020supervised} is a collection of real-world raw videos used for fine-tuning and testing.  It includes 6 scenes for training and 5 scenes for evaluation and testing. 
Each scene contains 7 frames with 5 different noise levels.
The ground truth frames are produced by taking the average of multiple noisy frames.

\noindent \textbf{Implementation details.}
Following previous works \cite{sheth2021unsupervised}, for natural video denoising, we adopt a batch size of 32 with $\ \mathcal{L}_{log}$ between the noisy and output videos to combine the 
Gaussian prior \cite{laine2019high,sheth2021unsupervised}. 
For raw video, we adopt a batch size of 8 with $\ \mathcal{L}_{2}$ loss.
The length of the input sequence is 9 for natural videos and 7 for raw videos.
To save training time for natural videos, we output and supervise the central frame and keep this setting while testing.
The learning rate starts with $1e$-4, where the Adam optimizer is adopted.
For raw dataset, overfitting can be severe since its size is extremely small. Though we 
get promising results with a simple way of fusion, \textit{i.e.}, blind convolutions in Distant Feature Fusion,
extra BSM does not yield satisfactory results, so we do not adopt BSM for 
raw dataset.
We conduct experiments in PyTorch 1.8.0 and
the Nvidia RTX 3090.

\noindent \textbf{Metrics.} 
Peak signal-to-noise ratio (PSNR) and structural similarity (SSIM) are the two metrics used for evaluation \cite{wang2004image}.
The larger value of these two metrics denotes better image quality.
We compare the performance of RDRF with several traditional video denoising methods, \textit{e.g.}, VNLB \cite{lebrun2013nonlocal}, VBM4D \cite{maggioni2012video}, supervised methods, \textit{e.g.}, VNLnet \cite{davy2019non}, DVDnet \cite{tassano2019dvdnet}, FastDVDnet \cite{tassano2020fastdvdnet}, FloRNN \cite{li2022unidirectional}, and unsupervised methods, \textit{e.g.}, MF2F \cite{dewil2021self}, UDVD \cite{sheth2021unsupervised}.

\begin{figure}[h]
\small
	   \begin{center}
		   \setlength{\tabcolsep}{0.8mm}
		   \renewcommand\arraystretch{1.6}
		   \begin{tabular}{cccc} 
   \includegraphics[height=0.23\linewidth]{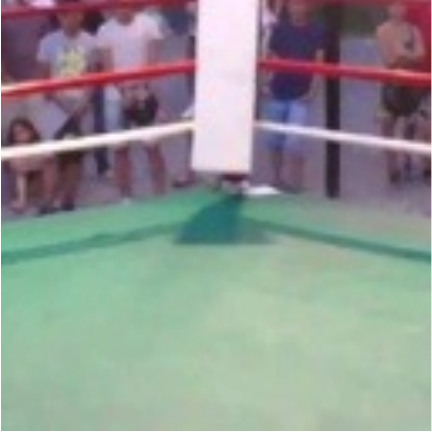}&
   \includegraphics[height=0.23\linewidth]{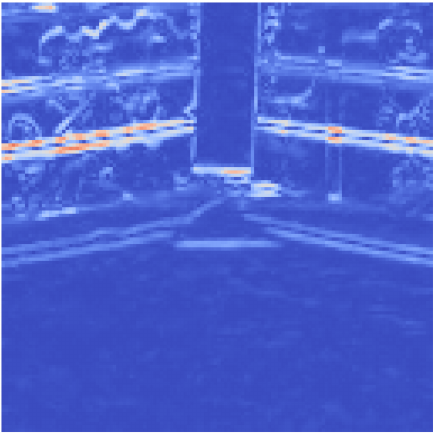}&	 
   \includegraphics[height=0.23\linewidth]{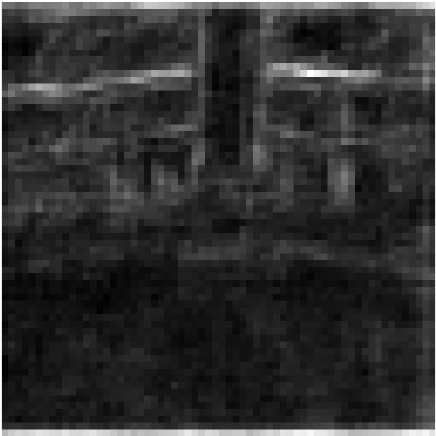}&
   \includegraphics[height=0.23\linewidth]{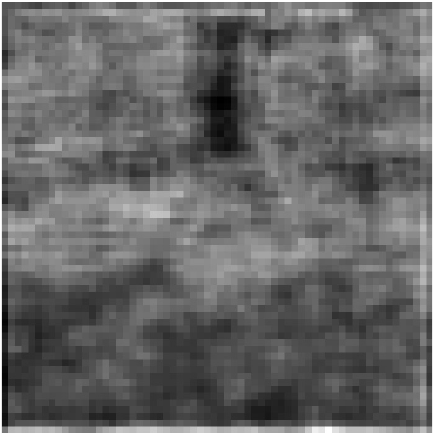}
   \\	
   \includegraphics[height=0.23\linewidth]{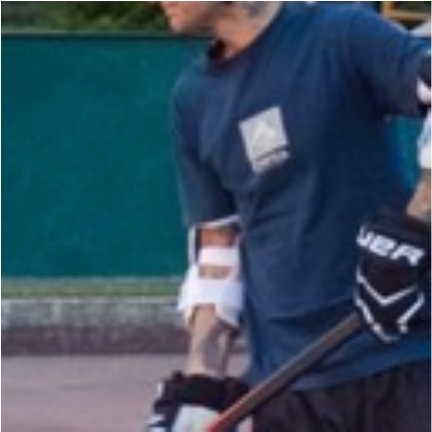}&
   \includegraphics[height=0.23\linewidth]{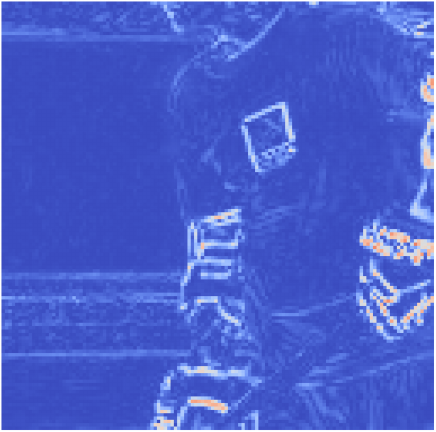}&
   \includegraphics[height=0.23\linewidth]{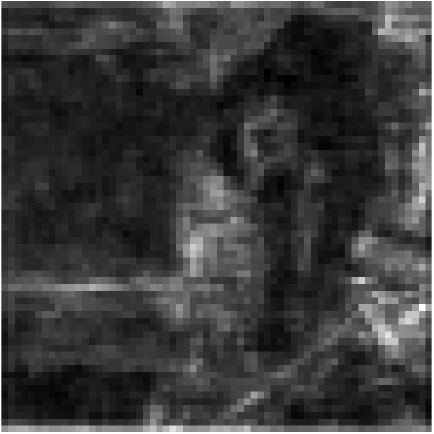}&
   \includegraphics[height=0.23\linewidth]{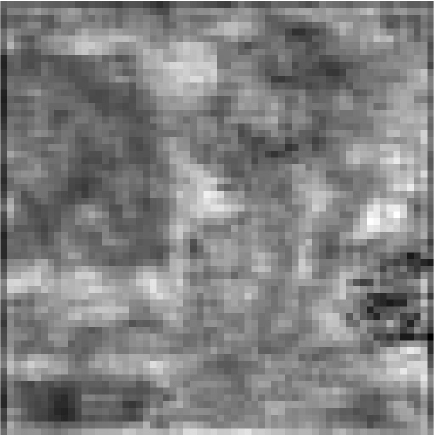}	
   \\		
   \includegraphics[height=0.23\linewidth]{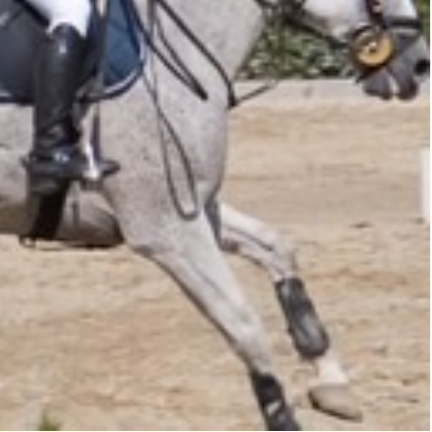}&
   \includegraphics[height=0.23\linewidth]{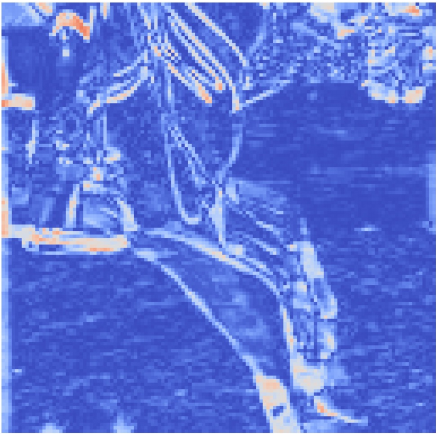}&
   \includegraphics[height=0.23\linewidth]{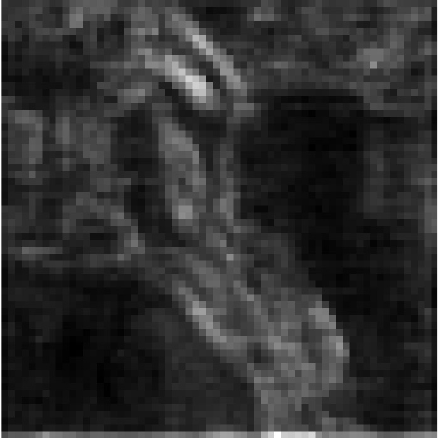}&		
   \includegraphics[height=0.23\linewidth]{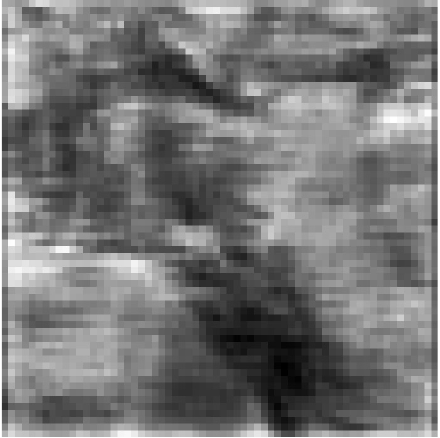}		
   \\
   Input frame & 
\makecell{Temporal \\difference map}
    & Beta & Gamma
   \\
		   \end{tabular}
		   \end{center}
     \vspace{-4mm}
		\caption{
  Illustrations on the pixel-wise difference between two consecutive frames, and corresponding learned modulation parameters, \textbf{i.e.}, $\gamma$ and $\beta$. }
  \vspace{-5mm}
		\label{fig:bsm}
   \end{figure}

\vspace{-2mm}
\subsection{Evaluation on Natural Videos}

We conduct extensive experiments to compare our method with three types of methods, including traditional, supervised and unsupervised methods. The results on the Set8 dataset and the Davis test set are reported in Table \ref{tab:res_set8} and Table \ref{tab:res_davis} respectively.

\begin{figure*}
 \centering
\hspace{-62.5mm}
 \begin{minipage}[t]{0.67\linewidth}
  \begin{tabular}{cccc}
  
  \includegraphics[width=0.35\linewidth]{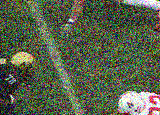}
  &
   \includegraphics[width=0.35\linewidth]{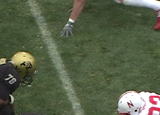} 
     & \includegraphics[width=0.35\linewidth]{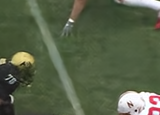}
  & \includegraphics[width=0.35\linewidth]{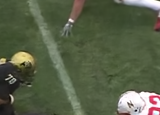}
     \\
   
    Noisy & Clean & DVDnet$^{\dagger}$ \cite{tassano2019dvdnet} & FastDVDnet$^{\dagger}$ \cite{tassano2020fastdvdnet} \\ 
    14.82/0.1180 & PSNR/SSIM & 31.46/0.7614 & 31.99/0.7935
    \\

   \includegraphics[width=0.35\linewidth]{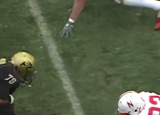}
   &
 \includegraphics[width=0.35\linewidth]{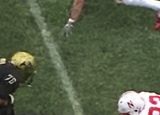}
  &\includegraphics[width=0.35\linewidth]{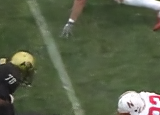}
  &\includegraphics[width=0.35\linewidth]{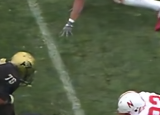}  
  \\ 

  FloRNN$^{\dagger}$ \cite{li2022unidirectional} & MF2F \cite{dewil2021self} &  UDVD \cite{sheth2021unsupervised} &  RDRF (Ours)  \\
    33.16/0.8345 &  30.26/0.7246 & 31.49/0.7730 &  32.75/0.8277
  \\
  \end{tabular}
 \end{minipage}

 \vspace{-3mm}
 \caption{Visual quality comparison on Set8 dataset. Supervised methods are marked with ${\dagger}$.}
    \vspace{-2mm}
 \label{fig:set8} 
\end{figure*}

\begin{figure*}
 \centering
\hspace{-62.5mm}
 \begin{minipage}[t]{0.67\linewidth}
  \begin{tabular}{cccc}
  
  \includegraphics[width=0.35\linewidth]{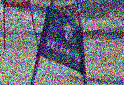}
  &
   \includegraphics[width=0.35\linewidth]{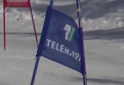} 
     & \includegraphics[width=0.35\linewidth]{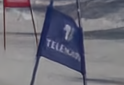}
  & \includegraphics[width=0.35\linewidth]{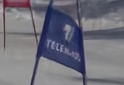}
     \\
   
    Noisy & Clean & DVDnet$^{\dagger}$ \cite{tassano2019dvdnet} & FastDVDnet$^{\dagger}$ \cite{tassano2020fastdvdnet} \\ 
    14.36/0.0640 & PSNR/SSIM & 36.07/0.9236 & 36.45/0.9307
    \\

   \includegraphics[width=0.35\linewidth]{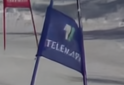}
   &
 \includegraphics[width=0.35\linewidth]{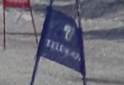}
  &\includegraphics[width=0.35\linewidth]{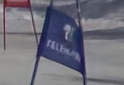}
  &\includegraphics[width=0.35\linewidth]{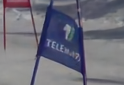}  
  \\ 

  FloRNN$^{\dagger}$ \cite{li2022unidirectional} & MF2F \cite{dewil2021self} &  UDVD \cite{sheth2021unsupervised} &  RDRF (Ours)  \\
    37.66/0.9403 &  33.46/0.8284 & 36.25/0.9274 &  37.45/0.9392
  \\
  \end{tabular}
 \end{minipage}

 \vspace{-3mm}
 \caption{Visual quality comparison on Davis dataset. Supervised methods are marked with ${\dagger}$.}
    \vspace{-3mm}
 \label{fig:davis} 
\end{figure*}

We obtain 
superior results in quantitative and qualitative metrics than previous unsupervised methods, where the qualitative results on Set8 and Davis benchmark datasets are shown in Figure \ref{fig:set8} and Figure \ref{fig:davis}, respectively. Our results show more detailed structures, also smoother in the flat areas.
For UDVD \cite{sheth2021unsupervised}, since the temporal neighbor pixels are masked from the receptive field, the limited useful information brings inferior performance.
For MF2F \cite{sheth2021unsupervised}, the performance is limited by the lack of information in the reference frame and the error by misalignment.
Instead, our method can fully exploit the spatial and temporal neighbors, also extracting the distant and near-temporal information. Thanks to these designs, our method outperforms previous unsupervised methods, which is even comparable to the supervised video denoising method.

\subsection{Evaluation on Real Raw Data}

\begin{figure*}[htbp]
\tabcolsep=1.2mm
  \centering
  \begin{tabular}{cc@{\hskip 0.16in}cc@{\hskip 0.16in}cc}
    
    \includegraphics[width=0.15\linewidth]{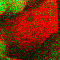} &
    \includegraphics[width=0.15\linewidth]{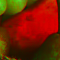} &
    \includegraphics[width=0.15\linewidth]{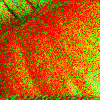} &
    \includegraphics[width=0.15\linewidth]{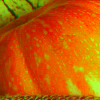}
    &
\includegraphics[width=0.15\linewidth]{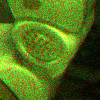} &
    \includegraphics[width=0.15\linewidth]{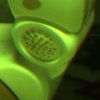}
     \\
    Noisy & Clean & Noisy & Clean & Noisy & Clean \\
    16.45/0.2224 & PSNR/SSIM & 14.32/0.1833 & PSNR/SSIM & 19.13/0.3614 & PSNR/SSIM\\
    
    \includegraphics[width=0.15\linewidth]{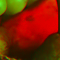} &
    \includegraphics[width=0.15\linewidth]{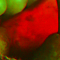} &
    \includegraphics[width=0.15\linewidth]{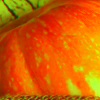} &
    \includegraphics[width=0.15\linewidth]{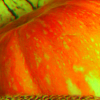}
    &
    \includegraphics[width=0.15\linewidth]{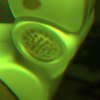} &
    \includegraphics[width=0.15\linewidth]{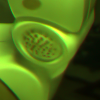}
     \\
    UDVD \cite{sheth2021unsupervised} & RDRF (Ours)  & UDVD \cite{sheth2021unsupervised} & RDRF (Ours)  & UDVD \cite{sheth2021unsupervised} & RDRF (Ours) 
    \\
    29.92/0.8866 & 31.45/0.9072 & 25.88/0.8108 & 27.43/0.8616 & 35.08/0.9370 & 35.26/0.9413
    \\
    
  \end{tabular}
  
  \vspace{-2mm}
\caption{Visual quality comparison on CRVD dataset \cite{yue2020supervised}.
}
 \vspace{-2mm}
\label{fig:crvd}

\end{figure*}

We evaluate the performance on the real world raw video denoising dataset \cite{yue2020supervised} in Table \ref{tab:crvd}. 
Our method outperforms UDVD \cite{sheth2021unsupervised} on almost all the noise levels due to our better utilization of reference and neighbor features. 
Furthermore, under this setting, other frame-based methods methods like MF2F \cite{dewil2021self} cannot be directly applied to raw videos, since their method relies on a pre-trained backbone that only works in the RGB color domain.
We also compare the performance of our RDRF with supervised methods RViDeNet \cite{yue2020supervised}. 
We achieve better quantitative performance than UDVD and RViDeNet at all noise levels. We emphasize that UDVD is directly trained on the mosaiced raw videos, and RViDeNet is pre-trained and fine-tuned on paired videos. 
In Figure \ref{fig:crvd}, we provide visualizations of restored frames on the CRVD dataset. Our method can better reconstruct the fine textures
by
the full exploitation of the reference and neighbor frames, also with the temporal similar patterns.

\subsection{Analysis of the Proposed Method}

\begin{figure}[t]
\centering
\includegraphics[width=1\linewidth]{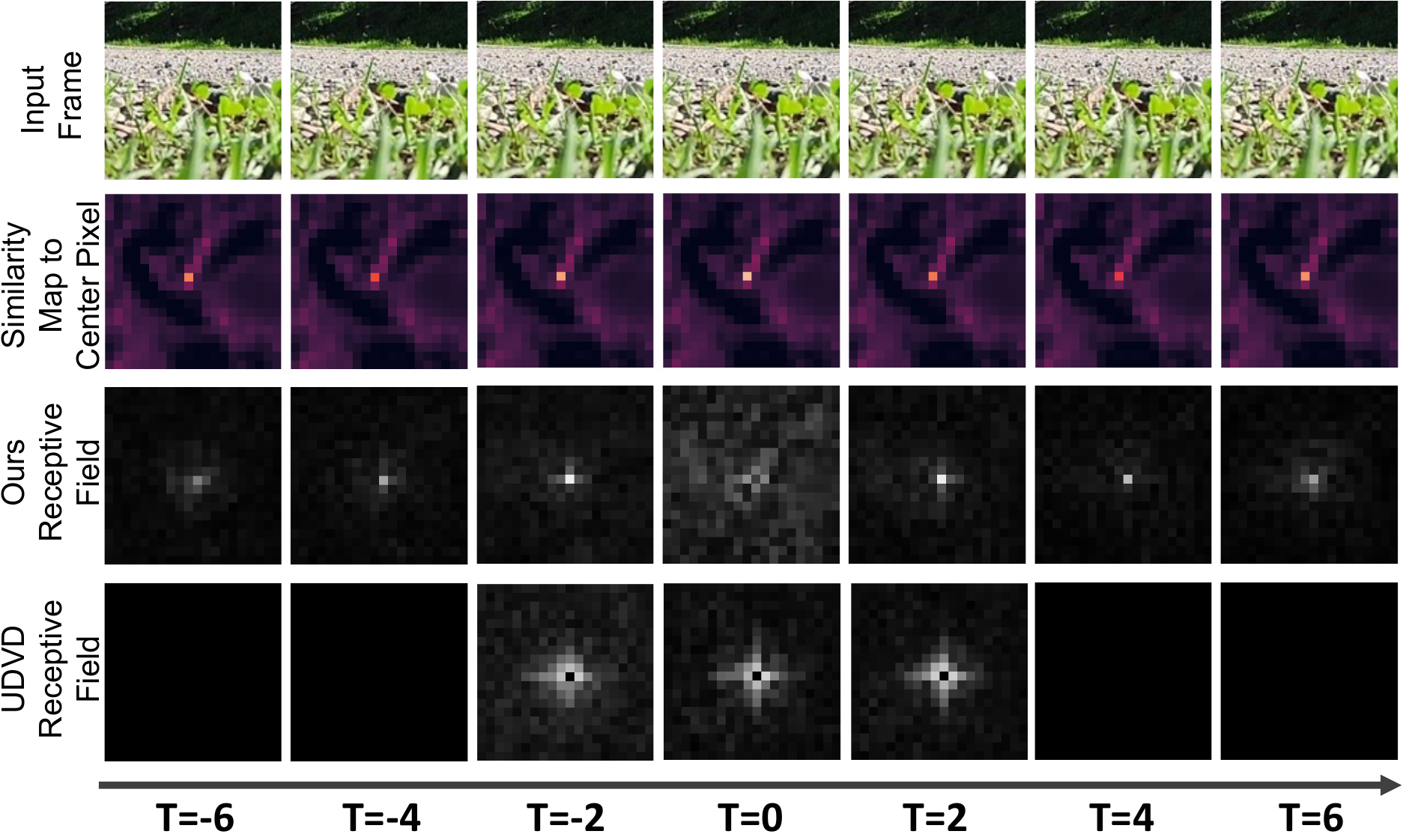}

\caption{
The degree of similarity to the center pixel of the reference frame (T=0) and the receptive field comparison. Our method can focus on the most similar positions, \textit{i.e.}, the center pixels of neighbor frames, with a long temporal range. While UDVD
\cite{sheth2021unsupervised} is unable to utilize long-term information, and pixels most focused on by UDVD may not contain the most useful information without the dense receptive field.
}
\vspace{-1mm}
\label{fig:rec}
\vspace{-3mm}
\end{figure}

\noindent\textbf{Visualization of Blind Spatial Modulation.}
The purpose of the Blind Spatial Modulation module is to fuse the current frame and its adjacent temporal features in a spatially adaptive way. We visualize the corresponding feature maps in Figure \ref{fig:bsm}. 
Regions with smaller motion
correspond to smaller beta and larger gamma values. This is
because when the adjacent frame has a higher degree of similarity,
it can be directly used for recovery
without much modification (smaller beta), and also being indispensable in the recovery process (larger gamma).
As a result, this allows for the effective utilization of temporal features, while implicitly capturing motion information.

\noindent\textbf{Visualization on receptive field.} To better validate the significance of a denser receptive field, we provide the corresponding visualization shown in Figure \ref{fig:rec}.
As can be seen, when the scene motion is small, the center pixel of the neighbor frame is most similar to the center pixel of the reference frame. 
This proves that the center pixels of neighbor frames can provide the most useful hints to restoring the reference frame, while UDVD \cite{wu2020unpaired} is unable to leverage this information, due to its under-utilization of the neighbor features. 
Also, the local aggregation also limits the range of available temporal information.
Our approach, on the other hand, can efficiently use both distant temporal information and the current frame. 
By focusing on the most similar patterns for the restoration, it enables the potential for better performance.

\noindent\textbf{Ablation studies on the proposed modules.} 
Our method consists of two branches for temporally long-range information aggregation and local neighbor information processing. For both the two parallel branches,
our method can fully exploit the features of the reference frame and its temporal neighbors. 
We validate that both of these branches are indispensable for self-supervised video denoising. 
We do ablation studies on the Set8 dataset, as
seen
in Table \ref{tab:ablation}.

\noindent \textbf{Ours w/o Distant Feature Fusion.} The lack of long-term features will cause a significant limitation on the scope of temporal information, where a large number of similarity patterns in the neighbor frames can not be utilized. This leads to a 1.41 dB drop in PSNR. 

\noindent \textbf{Ours w/ blind convolutions for propagation.}
To further validate that only gathering distant temporal features without the dense receptive field for neighbor frames is not enough, we implement $F_b$ and $F_f$ with blind convolutions following HQ-SSL \cite{laine2019high}. It can be found that  
full utilization of neighbor frames
boost recovery performance, proving
the importance of the denser receptive field.

\noindent \textbf{Ours w/o Local Feature Extraction.}
When relying solely on temporal propagation, 
the accumulated error during the propagation process will harm the performance of the network. The PSNR of the network decreases by 0.16 dB without Local Feature Extraction. 

\begin{table*}[h]
\tabcolsep=2.6mm
\begin{tabular}{cccccccc}

\toprule
Method & \multicolumn{5}{c}{Supervised}        & \multicolumn{2}{c}{Unsupervised} \\
ISO   & \multicolumn{1}{c}{FastDVDnet \cite{tassano2020fastdvdnet}} & RViDeNet \cite{yue2020supervised} & EDVR \cite{wang2019edvr} & LLRVD \cite{fu2022low} & FloRNN \cite{li2022unidirectional} & UDVD \cite{sheth2021unsupervised} & RDRF (Ours) \\
\cmidrule[0.4pt](lr){1-1}
\cmidrule[0.4pt](lr){2-6}
\cmidrule[0.4pt](lr){7-8}

    1600  & 43.43/0.9866 & 47.74/0.9938 & 47.59/0.9932 & 47.91/0.9941 & \uline{48.81/0.9956} & 48.02/0.9982 & \textbf{48.38/0.9983} \\
    3200  & 42.91/0.9844 & 45.91/0.9911 & 45.15/0.9889 & 46.02/0.9913 & \uline{47.05/0.9933} & 46.44/0.9980 & \textbf{46.86/0.9981} \\
    6400  & 40.29/0.9793 & 43.85/0.9880 & 43.10/0.9848 & 44.13/0.9885 & \uline{45.09/0.9910} & 44.74/0.9972 & \textbf{45.24/0.9975} \\
    12800 & 36.05/0.9613 & 41.20/0.9819 & 40.88/0.9779 & 41.58/0.9827 & \uline{42.63/0.9866} & 42.21/0.9966 & \textbf{42.72/0.9969} \\
    25600 & 36.50/0.9400 & 41.17/0.9821 & 41.03/0.9785 & 41.53/0.9829 & \uline{42.19/0.9872} & 42.13/0.9951 & \textbf{42.25/0.9948} \\
\midrule
avg   & 39.84/0.9703 & 43.97/0.9874 & 43.55/0.9847 & 44.23/0.9879 & \uline{45.15/0.9907} & 44.71/0.9970 & \textbf{45.09/0.9971} \\
\bottomrule
\end{tabular}%
  \caption{
  Performance comparison on CRVD \cite{yue2020supervised} dataset.
  The best results for un-/supervised methods are in bold and underlined.
  }
  \vspace{-4mm}
  \label{tab:crvd}
\end{table*}

\noindent \textbf{Ours w/o Blind Spatial Modulation.} 
Furthermore, the way to effectively fuse the reference and neighbor features is also the key to video restoration. 
Blind Spatial Modulation can explicitly consider the motion changes in the video and modulate the neighboring feature according to the motion in different regions. When we remove the Blind Spatial Modulation, it makes the network unable to dynamically perform  the fusion process, where a fixed fusion can lead to a decrease in performance.

\begin{table}[t]
\begin{tabular}{lcc}
\toprule
Method  & PSNR  & SSIM \\
\midrule
Ours w/o Distant Feature Fusion &   31.32    & 0.8563 \\
Ours w/ \ \ blind convolutions for propagation &  32.21    & 0.8865 \\
\midrule
Ours w/o Local Feature Extraction & 32.57 &  0.8933 \\
Ours w/o Blind Spatial Modulation &   32.52    & 0.8925 \\
Ours full  &   \textbf{32.73}    &  \textbf{0.8965} \\
\bottomrule
\end{tabular}%
  \caption{Ablation studies on the proposed method.}
  \vspace{-8mm}
  \label{tab:ablation}
\end{table}

\section{Conclusion}

In this paper, we present RDRF to tackle the challenges of self-supervised video denoising, including insufficient utilization of reference and neighbor information, also the limited available temporal feature. Our method consists of two branches for temporal long-range information aggregation and local neighbor information extraction. 
This enables our method fully exploit the information of spatial and temporal neighbors. 
Extensive results on both synthetic and real videos validate our superiority.

\vspace{1mm}
\noindent 
\textbf{Acknowledgement} This work was supported by the National Natural Science Foundation of China (62171038, 62171042, and 62088101), the R\&D Program of Beijing Municipal Education Commission (KZ202211417048), and the Fundamental Research Funds for the Central Universities.

\normalem
\bibliographystyle{ACM-Reference-Format}
\bibliography{sample-base}

\appendix

\end{document}